\pdfoutput=1
\documentclass[11pt]{article}

\usepackage[final]{acl}

\usepackage{times}
\usepackage{latexsym}

\usepackage[T1]{fontenc}
\usepackage[utf8]{inputenc}

\usepackage{microtype}

\usepackage{inconsolata}

\usepackage{graphicx}
\graphicspath{ {./images/} }

\usepackage{amsmath,amssymb,amsbsy}

\usepackage{subcaption}

\title{Reliability of Topic Modeling}

\author{Kayla Schroeder \\
  Department of Statistics \\
  Northwestern University \\
  \texttt{kaylaschroeder2026@u.northwestern.edu} \\\And
  Zach Wood-Doughty \\
  Department of Computer Science \\
  Northwestern University \\
  \texttt{zach@northwestern.edu} \\}

\begin{document}
\maketitle
\begin{abstract}

Topic models allow researchers to extract latent factors from text data and use those variables in downstream statistical analyses.
However, these methodologies can vary significantly due to initialization differences, randomness in sampling procedures, or noisy data.
Reliability of these methods is of particular concern as many researchers treat learned topic models as ground truth for subsequent analyses.
In this work, we show that the standard practice for quantifying topic model reliability fails to capture essential aspects of the variation in two widely-used topic models.
Drawing from a extensive literature on measurement theory, we provide empirical and theoretical analyses of three other metrics for evaluating the reliability of topic models.
On synthetic and real-world data, we show that McDonald's $\omega$ provides the best encapsulation of reliability.
This metric provides an essential tool for validation of topic model methodologies that should be a standard component of any
topic model-based research.  

\end{abstract}

\section{Introduction}

Over twenty years after its popularization by \citet{blei2003latent}, topic modeling remains a leading methodology for analyzing natural language.
Researchers use models such as Latent Dirichlet Allocation (LDA) and its many variations to uncover latent topics in human language, with applications  ranging from mathematics to genomics \citep{bravo2019cistopic, poushneh2022can, kukreja2023recent, yu2023discovering}.
These methods convert high-dimensional data into low-dimensional topic representations which are more easily used in downstream analyses \citep{gentzkow2019text}.
When it comes to human language and similar domains, topic models clearly cannot capture the full complexity of the data, so they instead provide a useful approximation of the data’s underlying structure \citep{doogan2021topic}.
LDA and many topic models are fit with approximate methods due to their intractable likelihoods; these implementations may be sensitive to random initialization, hyperparameter selection, or small perturbations to the data \citep{ramirez2012topic,lau2016sensitivity}.
We analyze topic models through the lens of internal consistency reliability, offering insight into the downstream analyses which they enable. Internal consistency reliability, a well-established concept in psychometrics, describes how well varying items measure the true underlying construct \citep{cronbach1951coefficient}. In this setting, reliability measures the extent to which different replications of a topic model consistently capture the underlying thematic structure. While there may be multiple suitable granular structures for different tasks, our focus is on identifying the optimal structure that best represents the underlying organization within a specific dataset.

We conceptualize topic modeling as a measurement problem; the lower-dimensional representation produced by a topic model provides an imperfect measure of the data’s underlying structure.
Many analyses relying on topic models do not consider the reliability of these measurements nor the effect that unreliability could have on their analyses.
\citet{yu2023discovering}, for example, train an LDA topic model on research abstracts to quantify trends in AI research; if repeated re-trainings of LDA produced significant variation in the learned topics, it would undermine any research conclusions drawn from those topics.
This is true for many other applications of topic modeling, from analyses of consumer complaints, academic peer review, and Twitter posts \cite{bastani2019latent,poushneh2022can,xue2020public}.
A better understanding of topic modeling reliability can strengthen our understanding of these papers and many like them.

Reliability is not a new consideration for the NLP field. Inter-annotator agreement (IAA) has been widely studied in analyses of human labelers \citep{bhowmick2008agreement,nowak2010reliable,amidei2019agreement}.
IAA has also been explored during the interpretation of topic models; 
\citet{prollochs2020business} analyzes reliability of annotators' ability to name topics from their most prevalent words.
Compared to this widespread study of the reliability of annotators, there has been relatively little exploration of the reliability of topic models themselves.
\citet{rudiger2022topic} highlights in particular the need for a measure of the \emph{internal consistency} of such models.

The study of internal consistency reliability was first popularized by \citet{cronbach1951coefficient} but has been widely studied in the statistical literature \citep{mcdonald2013test,tavakol2011making,kamata2003estimating,li1996reliability}.
Despite the existence of this statistical theory, the widely-adopted standard practice for measuring topic model reliability is an ad hoc approach \cite{maier2018applying}.
This approach repeatedly fits a topic model using different random initializations and calculates the proportion of topic matches.
Here, two topics are said to match if the cosine similarity between their top-word probabilities is above a threshold of 0.7.
Among the issues with this method is its use of a fixed threshold, which has been criticized in studies of IAA \citep{reidsma2008reliability}. 

In this paper, we show that the lack of theoretical grounding and omission of the document over topics distribution in the standard practice method results in an entirely insufficient reliability quantification.  
Building on the statistical literature, we introduce and evaluate methods that 
encompass both the reliability of topic models in terms of top words and distribution of documents over topics. \footnote{The code used for the analyses and metrics can be accessed at \href{https://github.com/kaylaschroeder/reliability}{https://github.com/kaylaschroeder/reliability}.}

\section{Related Work}
\label{sec:related_work}

Most work related to topic model reliability has been focused on topic model quality. Coherence, a measure of quality and human interpretability of a topic, has been widely studied \citep{doogan2021topic, hoyle2021automated}. Topic model quality differs fundamentally from topic model reliability.

Existing methods attempt to quantify topic model reliability or stability either employ variations of similarity measures and/or require domain knowledge.
\citet{chuang2015topiccheck} quantifies the stability of topic models using similarity and domain specific information, differing from our approach both in methodology and in their requirement for domain specific knowledge. Given its reliance on domain knowledge, this method is unrelated to our work and not directly comparable. \citet{rieger2024ldaprototype} proposes assessing reliability of the topics and the topic model using varying similarity measures. \citet{ballester2022robustness} proposes the asymptotic average standard deviation of pairwise similarity as a measure of robustness, and argues that robustness is akin to reliability. 
\citet{maier2018applying} proposes a reliability measure comprised of the proportion of top-word cosine similarities above a threshold of 0.7 and is currently the widely accepted standard practice for reliability of topic models. 
These methods for quantifying reliability are not rooted in statistical practice, an absence shown to be problematic for inter-annotator reliability in \citet{reidsma2008reliability}. We emphasize that the similarity measure is not a measure of reliability and showcase the shortcomings of this type of quantification.

Internal consistency reliability has specifically been called for in embeddings, in scores for LLM performance on benchmarks, and in NLP methods at large \citep{du2021assessing, xiao2023evaluating, riezler2022validity}. Further NLP literature uses the term reliability, but does not refer to statistical reliability measures \citep{tan2021reliability, elder2020make, rios2020empirical, dunn2022predicting}.

\section{Preliminaries: Unidimensional Reliability}

Reliability literature focuses heavily on unidimensonal cases \citep{cronbach1951coefficient, mcdonald2013test}. We first consider unidimensional reliability to investigate assumptions for the multidimensional counterparts. Unidimensional reliability is later employed to quantify reliability of a trivial topic model with two topics, as only one topic is necessary for identifiability. 

\subsection{Cronbach's Alpha}

Cronbach's alpha has historically been the most widely used and accepted statistical method for objective reliability estimation \citep{tavakol2011making}. This measure of internal consistency developed within the statistical literature to address the shortcomings of the split-half coefficients, a method that splits the data in half and compares each half to obtain a reliability measure \citep{cronbach1951coefficient}. The formula for Cronbach's alpha is 

\begin{equation}
  {\alpha = \frac{n\cdot \bar{c}}{\bar{v}+(n-1)\cdot \bar{c}}} \label{eq:cron_alph} 
\end{equation}

\noindent where $n$ is the number of replications, $\bar{v}$ is the average replication-specific variance, and $\bar{c}$ is the average inter-replication covariance. 

The potential shortcomings of Cronbach's alpha in the topic model setting stem largely from the method's assumptions. To satisfy the assumptions of Cronbach's alpha, values are assumed to (1) be continuous and normally distributed, (2) have additive measurement error, have uncorrelated errors, (3) measure a single latent trait, and (4) have the same relationship with the underlying construct (or topic in this setting) \citep{tavakol2011making}. Even in the trivial topic modeling setting, the assumptions of additive measurement error and normality are unlikely to hold. Some prior work in \citet{xiao2023performance} suggests that Cronbach's alpha can withstand less severely non-normal data, however a topic's distribution over documents are severely non-normal and thus this does not apply. 

\subsection{McDonald's Omega}

McDonald's omega is the most popularized alternative to Cronbach's alpha \citep{hayes2020use}. Seeking to relax the requirement of all items having the same relationship with the underlying construct (an assumption termed tau-equivalence), McDonald's alpha is formulated as 

\begin{equation}
    \omega = \frac{(\sum_{i=1}^{n}\lambda_i)^2}{(\sum_{i=1}^{n}\lambda_i)^2 + \sum_{i=1}^{n}\theta_{ii}}
\end{equation}

where $\lambda_i$ is the factor loading of the $i$th replication, $\theta_{ii}$ is the error variance of the $i$th replication, and $n$ is the number of replications \citep{mcdonald2013test}. Each factor loading represents the variance of the unknown topic distribution across documents that is explained by each individual replication and is given by the correlation between the replication and the unknown underlying topic. Factor analysis, a common multidimensional methodology, is employed to obtain the factor loadings using the formulation 
\begin{equation}
    \mathbf{X} = \mathbf{M} + \mathbf{L}F + \mathbf{\epsilon}
\end{equation}

with observation matrix $\mathbf{X}$, factor loading matrix $\mathbf{L}$, factor $F$, and error term matrix $\mathbf{\epsilon}$ \citep{lawley1962factor}. In our setting, $\mathbf{X}$ is a matrix with columns as topic's distribution over documents from each replication and $F$ represents an unknown vector of the underlying distribution over documents for a singular topic.

The formulation of omega itself encompasses a larger family of reliability estimates, of which Cronbach's alpha is a restrictive case, in which the requirements for uncorrelated errors, normality, tau-equivalence and unidimensionality are not required \citep{viladrich2017journey}. Given that different replications of topic models have different initializations but the same underlying algorithm, tau-equivalence and uncorrelated errors are expected to be upheld. Normality, however, is not likely to be upheld. McDonald's omega, then, is expected to provide improvements upon Cronbach's alpha. 

\subsection{Spearman-Brown Reliability}

The Spearman-Brown equation is a broadly defined reliability measure that assesses the impact of additional test items (in our setting, replications) to obtain an internal consistency reliability score \citep{walker1953statistical}. This equation is defined

\begin{equation}
        R = \frac{nr}{1+(n-1)r}
        \label{eq:sb_reliab}
\end{equation}

\noindent for $n$ replications and correlation $r$. Equal correlation and variance between items in addition to additive (linear) measurement error comprise the key assumptions of this measurement, rendering Spearman-Brown reliability much more flexible in wide-ranging settings. Assumptions of equal correlation and variance follow easily from the independence of replications. While the Spearman-Brown equation does still suffer from the assumption of linearity in errors when applied to topic modeling, we note that normality is not required.

\section{Topic Model Reliability Methods}

In the context of nontrivial topic modeling, where numerous topics are considered, unidimensional measurements or groups thereof are inadequate representations of the data. We propose three multidimensional counterparts as potential suitable extensions for topic model reliability.

To adequately compare topics, topics must be aligned first to determine their counterpart in other replications. We use the standard practice topic matching method defined in \citet{maier2018applying} of matching topics with the highest cosine similarity among top-word probabilities. This matching choice is selected purposefully to allow for a direct comparison between the current standard practice reliability measure from \citet{maier2018applying} and the reliability measures we introduce below.

\subsection{Stratified Alpha Coefficient}

The stratified alpha coefficient is the multidimensional extension of Cronbach's alpha and is defined

\begin{equation}
    \alpha_{str} = 1 - \frac{\sum_{i=1}^k \sigma_i^2 (1-\alpha_i)}{\sigma^2}
    \label{eq:alph_strat} 
\end{equation}

\noindent where $\alpha_i$ is the reliability of item $i$ defined by (\ref{eq:cron_alph}), $\sigma_i^2$ is the item variance and $\sigma^2$ is the overall variance of the test \citep{cronbach1965alpha}. In the topic model setting, the items $1,...,k$ refer to the individual topics (with one topic removed for identifiability purposes) and $k$ refers to the total number of topics. The stratum, or components of the test as they are described in \citet{cronbach1965alpha}, are then the topics themselves. Multidimensional measurement relaxes the unidimensionality assumption of Cronbach's alpha while retaining its other assumptions. Nonlinear errors and non-normality remain challenges within the classical representation of the stratified alpha coefficient.

Pooled estimates have a rich history in reliability quantification within meta-analysis, with applications spanning medicine to psychiatry \citep{bobos2020measurement,trajkovic2011reliability}. We adopt this technique to quantify the reliability of each topic (unidimensional alpha) by pooling variance and covariance estimates derived from both the distribution of documents over topics and the term distribution for each topic. Incorporating the distribution of documents over topics is crucial in topic model reliability as it plays a pivotal role in interpretation and prediction, as exemplified by \citep{bravo2019cistopic} where topic proportions are used to predict transcription factors. Our implementation utilizes \texttt{ltm} R package's Cronbach's alpha function \citep{rizopoulos2007ltm}. 

\subsection{Multivariate Omega}

Multivariate omega follows simply from the unidimensional McDonald's omega as omega does not require unidimensionality \citep{kamata2003estimating}. Relying on the same assumptions as McDonald's omega, McDonald's omega in the multivariate setting is defined
\begin{equation}
    \omega = \frac{\mathbf{1'}_n\boldsymbol{\lambda\lambda}' \mathbf{1}_n}{\sigma^2_X} .
\end{equation}

Matrix $\boldsymbol{\lambda}$ contains the aforementioned factor loadings and, in practice, $\sigma^2_X$ is the sum of the sample variance-covariance matrix elements for observation matrix $X$. 

As with Cronbach's alpha, we pool the omega estimates to obtain multivariate omega quantifying both the distribution of documents over topics and the posterior distribution of terms for each topic. Our implementation utilizes the \texttt{psych} R package's omega function \citep{revelle2015package}.

\subsection{Maximal Reliability}
\label{sec:max_reliab}

Maximal reliability extends the Spearman-Brown equation to multiple items by summing each item's contribution \citep{li1996reliability}. This formulation is 

\begin{equation}
    R_k = \frac{\frac{n_1r_1}{1-r_1}+\frac{n_2r_2}{1-r_2}+\ldots+\frac{n_Kr_K}{1-r_K}}{\frac{K}{1+(K-1)\rho}+\frac{n_1r_1}{1-r_1}+\frac{n_2r_2}{1-r_2}+\ldots+\frac{n_Kr_K}{1-r_K}}
    \label{eq:max_reliab} 
\end{equation}

\noindent where, for $i \in 1,\ldots,K$, $r_i$ is the topic reliability given by (\ref{eq:sb_reliab}) for each of the $K$ topics, $n_i$ is the number of replications, and $\rho$ is the common correlation between any two topics. 

Topic models are unlikely to uphold the linear relationship between two topics from differing replications, so the correlation coefficient should be chosen appropriately to characterize the true relationship between replications. For example, the Pearson correlation measure is 

\begin{equation}
    r = \frac{\sum_i(x_i-\bar{x})(y_i-\bar{y})}{\sqrt{\sum_i(x_i-\bar{x})^2\sum_i(y_i-\bar{y})^2}}
\end{equation}

for two vectors $\mathbf{x}$ and $\mathbf{y}$ focuses on the linear relationship between two vectors, and thus would not be appropriate in this context \citep{schober2018correlation}. Cosine similarity is defined 

\begin{equation}
    cos(\theta) = \frac{\sum_ix_i y_i}{\sqrt{\sum_ix_i^2}\sqrt{\sum_iy_i^2}}
\end{equation}

\noindent for two vectors $\mathbf{x}$ and $\mathbf{y}$.

It is evident from the formulation of correlation that correlation is the cosine similarity between centered vectors. Both cosine similarity and Pearson's correlation are traditional similarity measures \citep{verma2020new}. We also note the remarkable similarity between the formulation of Cosine similarity and the general correlation coefficient derived in \citet{kendall1948rank}. To adequately encompass the nature of the topic relations and the maximal reliability measurement, then, it is natural to instead use cosine similarity in place of the correlation coefficient. \citet{schmidt2014methods} recommends that two averaged correlations can contribute the singular value representing the study correlation. We apply the same methodology here to obtain a composite score using the averaging of the cosine similarity measures.

\section{Data}

To demonstrate the reliability of LDA in a controlled setting, we conducted experiments using both trivial and nontrivial synthetic datasets, each consisting of 10,000 texts generated using the framework outlined in \citet{wood2021generating}. The trivial dataset is a bag-of-words sample with a vocabulary of 16 and a ground truth of 2 topics. The nontrivial dataset is derived from a bag-of-words LDA model with 200 topics, trained on filtered webtext data \citep{radford2019language}. For the nontrivial data, we considered topic models with a misspecified number of topics (20, 50, and 100), simulating a common scenario in practice as determining the optimal number of topics remains a challenging task \citep{hasan2021normalized,zhao2015heuristic}. Considering reliability across models with varying numbers of topics remains crucial for applications that interpret topic models, such as \citet{yu2023discovering}, to provide a measure of the certainty in the research conclusions drawn from interpretations of topics.

We also applied our methodology to a real-world dataset of 5,223 consumer complaints about cryptocurrency companies filed with the CFPB, a corpus publicly available on the entity's website \citep{cfpb_dataset}. In this dataset, the true number of topics is unknown, so we investigated reliability variation across 20, 50 and 100 topics.

To assess reliability, we generated 100 replications for the application and nontrivial synthetic data and 10 replications for the trivial synthetic data, each with a randomly generated seed. This approach allows evaluation of the consistency of topic model results across various random initializations.

For synthetic and application data, topic model replications were generated using the \texttt{topicmodels} package from \citet{grun2011topicmodels} and the \texttt{stm} package from \citet{roberts2019stm}. Both packages employed a specified number of topics and their default settings, with Gibbs sampling for \texttt{topicmodels} and spectral decomposition (deterministic) for \texttt{stm}. These replications were executed on a CPU.

\section{Results}

Our three proposed reliability measures are compared with the widely accepted and adopted standard practice method from \citet{maier2018applying}. Despite the existence of other baselines such as those proposed by \citet{rieger2021rollinglda} and \citet{ballester2022robustness}, this work focuses on the current standard practice given that existing comparable baselines suffer from the same shortcomings as the standard practice due to the employment of similarity measures which are not a measure of reliability, as we discuss in Section \ref{sec:related_work}. While exploring alternative baselines is a worthwhile direction for future research, our choice to focus on the standard practice method is justified given its widespread adoption and the similar limitations of other available approaches. 

The usage of 0.7 as a cutoff within the current standard practice methodology is likely due to  existing literature that argues Cronbach's alpha should be $> 0.7$ to be considered reliable \citep{johnson2017reliability}.  While our results will show the detriment of the current practice metric's incorporation of such a hard cutoff with further investigation provided in Appendix \ref{sec:appendix_std_prac_cutoff}, related reliability literature is useful for interpreting and understanding resulting reliability measures. We use the widely accepted Cronbach's alpha rule of thumb, described in Table \ref{tab:rule_thumb}, for interpreting our reliability values as this rule is applied in practice to all types of reliability measures \citep{gliem2003calculating}.

\begin{table}[!th]
 \centering
\begin{tabular}{c|c}
    Cronbach's Alpha & Interpretation \\
   \hline
    $\alpha > 0.9$ & Excellent\\
    $0.9 > \alpha > 0.8$ & Good \\
    $0.8 > \alpha > 0.7$ & Acceptable \\
    $0.7 > \alpha > 0.6$ & Questionable \\
    $0.6 > \alpha > 0.5$ & Poor \\
    $\alpha < 0.5$ & Unacceptable 
\end{tabular}
\caption{Rule of thumb for interpreting Cronbach's alpha (and reliability measures as a whole).}
\label{tab:rule_thumb}
\end{table}

\subsection{Trivial Synthetic Data}

To illustrate the properties of our reliability metrics, we begin with a unidimensional analysis using the trivial setting. We examine a single topic in Table \ref{tab:triv}, as the rows of the document over topic distribution sum to 1. In Table \ref{table:triv_word_dist}, nine replications exhibit identical document over topic distributions, but replication 6 demonstrates a nearly equal balance of topic proportions within each document. Table \ref{table:triv_comp_table} presents a similar scenario, demonstrating that the posterior distribution of words over topics is consistent across 9 replications but deviates significantly in replication 6. This deviation leads to the misclassification of 9 documents in replication 6, reinforcing the notion that a single replication can be insufficient for characterizing a corpus. Reliability is essential to accurately assessing topic model results.

\begin{table} 
\begin{subtable}{.45\textwidth}
\centering
\begin{tabular}{c|c|c|c|c}
 & \textbf{Doc 1} & \textbf{Doc 2}  & \textbf{Doc 3} & \textbf{Doc 4}\\
 \hline
\textbf{Rep 5} & 0.002 & 0.998 & 0.998 & 0.002\\ %& 0.998 \\
\hline
\textbf{Rep 6} & 0.496 & 0.505 & 0.506 & 0.495\\ %& 0.503 
\end{tabular}
\caption{A snippet of the document over topic distribution. Replication 5 is nearly identical to all other replications not listed in the table. Replication 6 proportions are all within .01 of 0.5.}
\label{table:triv_word_dist} 
\end{subtable}

\begin{subtable}{.45\textwidth}
\centering
\begin{tabular}{c|c|c}
 & \textbf{Words} & \textbf{Docs}  \\
 \hline
\textbf{Rep 5} & a, d, e, g, h, j, m, p & 5066 \\ 
\hline
\textbf{Rep 6} & a, c, d, e, g, h, j, l, o, p & 5069 \\  
\end{tabular}
\caption{Comparison of replication 6 performance to all other replications. The table contains the words and number of documents (out of 10,000) classified into topic 1 (instead of topic 2) by each replication. While the document count only differs by 3, we note that replication 6 misclassifies 9 documents.}
\label{table:triv_comp_table} 
\end{subtable}

\begin{subtable}{.45\textwidth}
\centering
\begin{tabular}{c|c|c|c|c}
   \textbf{Type} &  \textbf{R} & $\boldsymbol{\alpha}$ & $\boldsymbol{\omega}$ & \textbf{Current} \\
   \hline
  \textbf{Full} & 0.995 & 0.991 & 0.997 & 1 \\
   % & (0.046) & (0.0004) & (0.0002) & (NA) \\
  \hline
  \textbf{Subset} & 0.856 & 0.420 & 0.913 & 1
\end{tabular}
\caption{Reliability of all replications and a subset of 2 replications. The current standard practice (`Current') is compared to the unidimensional versions of Maximal Reliability (`R'), Stratified Alpha ($\alpha$), and Multivariate Omega ($\omega$).}
\label{tab:triv_comp}
\end{subtable}
\caption{Trivial synthetic data reliability using topic 1.}
\label{tab:triv}
\end{table}

Table \ref{tab:triv_comp} describes the (unidimensional) reliability of the entire set of 10 replications and a subset of replications comprised of only replications 5 and 6, a setting which may be more realistic in practice due to LDA's computational intensity.\footnote{For the full analysis, the respective standard errors for
$\alpha$, $\omega$, and Spearman-Brown are
0.0004, 0.0002, and 0.005; for the subset analysis, they are 0.014, $7.2\mathrm{e}{-17}$, and 0.129 respectively. We have no way of calculating the standard practice method’s variance given the method’s lack of statistical grounding.} 
Clearly, the standard practice method fails to encompass variability in the document over topic distribution given its claims of perfect reliability among the replications. This contrasts with the unidimensional measures of maximal reliability, alpha, and omega which all effectively convey this existing variability among replications.

\subsection{Nontrivial Synthetic Data}

\begin{figure}[t]
\centering
\includegraphics[width=\columnwidth]{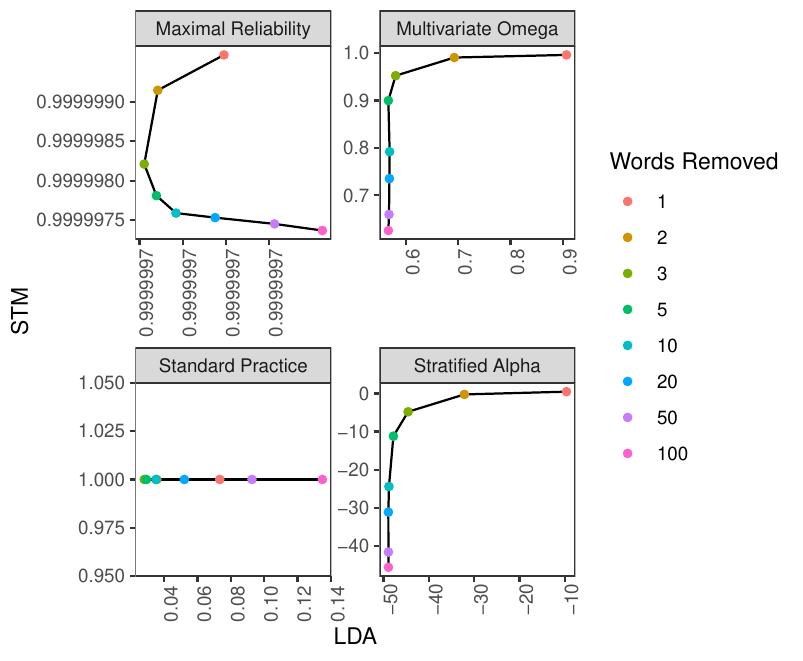}
\caption{Performance of individual reliability measures across varying numbers of random vocabulary word removals of \texttt{stm} on LDA.}
\label{fig:individ_comp_stm}
\end{figure}

To further compare performance of reliability measures, we examined both LDA and \texttt{stm} (Structural Topic Models). Introduced by \citet{roberts2019stm}, \texttt{stm} is a variant of LDA that incorporates covariates. Unlike LDA, \texttt{stm}'s initialization is deterministic, making it a suitable benchmark for comparing against LDA across replications. While \texttt{stm} remains consistent across different random seeds, any corpus of text should be considered a random draw from the broader distribution. Thus, perfect internal consistency of \texttt{stm} obscures the underlying randomness of the data and ``fixed randomness'' \citep{hellrich2016bad}.

Varying numbers of words are removed at random from the corpus before developing the \texttt{stm} and LDA models, akin to the practice of item-deletion which has previously been used to investigate reliability measures \citep{kopalle1997alpha}. While removing a few words from a large vocabulary has a minimal impact, removing a substantial number (e.g., 50 or 100) can significantly alter the corpus. Our findings confirmed this: with few word removals, topic models exhibited consistent top words, defined as the most probable terms within each topic, across replications, whereas more extensive removals led to deviations (see Appendix \ref{sec:appendix_mod_selction}). These results align with prior research on the influence of word frequency on topic similarity, emphasizing the importance of corpus composition for robust topic modeling \citep{rieger2021rollinglda}. A strong reliability measure should reflect such information.

\begin{figure}[t]
\centering
\includegraphics[width=\columnwidth]{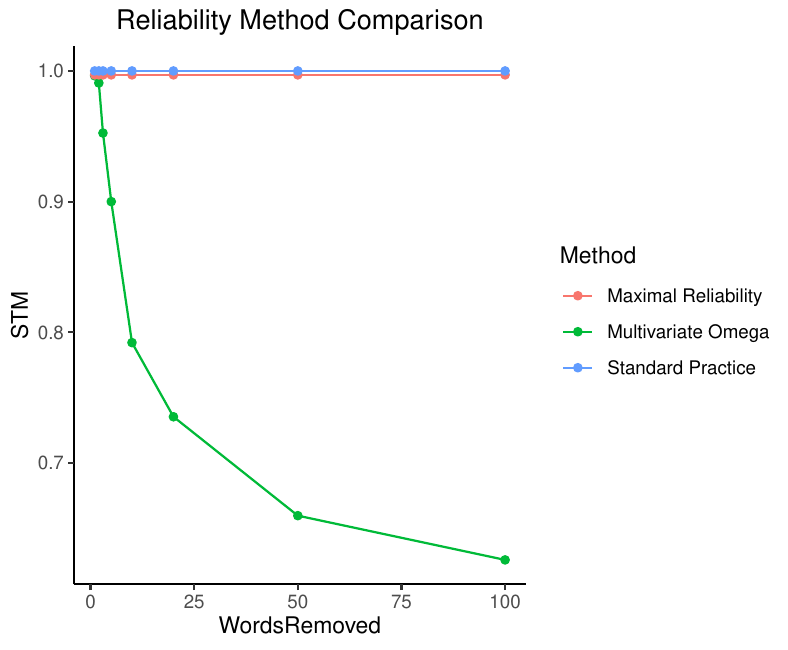}
\caption{Comparison of reliability methods of \texttt{stm} on increasing numbers of random vocabulary word removals. Note the nearly overlapping Maximal Reliability and Standard Practice results.}
\label{fig:method_comp_stm}
\end{figure}

Figure \ref{fig:individ_comp_stm} contrasts the effectiveness of various reliability methods against the standard practice comparing \texttt{stm} vs LDA models, controlling for initialization seed and word removal. The standard practice method proves inadequate, consistently overestimating reliability for \texttt{stm} models and underestimating it for LDA models. 
Stratified alpha yields exclusively negative values, suggesting a methodological flaw likely stemming from violated normality assumptions.
Maximal reliability lacks sensitivity and, while partially capturing the impact of word removal, exhibits an unexpected decline in reliability with fewer word removals, indicating poor quantification. In contrast, multivariate omega effectively reflects the impact of word removal on \texttt{stm} and LDA models, as further explored in Figure \ref{fig:method_comp_stm}. The deterministic nature of \texttt{stm} and the sensitivity of topic models to word removal necessitate a corresponding decrease in reliability with increasing word removals.
Our results validate this expectation, with multivariate omega demonstrating superior sensitivity to changes in word removal, making it a more suitable metric for quantifying topic model reliability.

To reinforce our conclusion, we analyzed the frequent and exclusive (FREX) words for the most prevalent topic across replications with varying levels of word removal. Consistency was observed with one word removal but not with 100 (Appendix \ref{sec:appendix_mod_selction}).

\begin{table}[t]
\begin{subtable}{.45\textwidth}
\centering
    \begin{tabular}{c|c|c|c}
         $\mathbf{R_k}$ & $\boldsymbol{\alpha_{str}}$ & $\boldsymbol{\omega}$ & \textbf{Current} \\
         \hline
         0.9999997 & -48.55 & 0.567 &  0.022
    \end{tabular}
    \caption{Reliability of LDA models with 100 topics comparing current standard practice (`Current'), Maximal Reliability ($\mathbf{R_k}$), Stratified Alpha ($\boldsymbol{\alpha_{str}}$), and Multivariate Omega ($\boldsymbol{\omega}$).}
    \label{tab:synth_n100_reliab}
 \end{subtable}
\begin{subtable}{.45\textwidth}
    \centering
    \begin{tabular}{c|c|c|c}
       \textbf{Topics} & \textbf{20} & \textbf{50} & \textbf{100} \\
       \hline
       \textbf{$\boldsymbol{\omega}$} & 0.693 & 0.694 & 0.567 \\
       % \hline
       (SE) & (0.010) & (0.006) & (0.004) \\
    \end{tabular}
    \caption{LDA reliability as given by Multivariate Omega ($\boldsymbol{\omega}$) for topic model replications with 20, 50, and 100 topics.}
    \label{tab:synth_omega_reliab}
\end{subtable}
\caption{Nontrivial synthetic data reliability across 100 replications.}
\label{tab:nontriv_reliab_res}
\end{table}

Our analysis underscores the limitations of the standard practice, and shortcomings of the proposed stratified alpha and maximal reliability. As illustrated by the reliability of synthetic data LDA models with 100 topics in Table \ref{tab:synth_n100_reliab}, these metrics can lead to drastically different assessments, potentially misleading researchers.\footnote{The standard errors of $\alpha_{str}$ and $\alpha_{str}$ are 0.002 and 0.004, respectively. We have no way of calculating the standard practice method's variance given the method's lack of statistical grounding. \citet{raykov2006direct} suggests an approximation for the standard error of ${R_k}$; we leave this for future work.} To address these shortcomings, we advocate for multivariate omega reliability as a more comprehensive and effective measure.

Table \ref{tab:synth_omega_reliab} further highlights the significant reliability challenges inherent in topic models, particularly as the number of topics increases. According to the Rule of Thumb in Table \ref{tab:rule_thumb}, multivariate omega reliability falls into the `questionable' category for 20 and 50 topics and the `poor' category for 100 topics. These low reliability values, solely due to random chance, cast doubt on the common practice of treating obtained topic models as definitive representations of underlying latent structures.

\subsection{Application}

\begin{table}[t]
    \centering
    \begin{tabular}{c|c|c|c}
         & \textbf{20 Topics} & \textbf{50 Topics} & \textbf{100 Topics} \\
         \hline
         % out of date: need full results
        $\boldsymbol{\omega}$ & 0.92283 & 0.82235 & 0.74957 \\
        (SE) & (0.05154) & (0.02104) & (0.01052) \\
    \end{tabular}
    \caption{Reliability (multivariate omega) and standard error of LDA models of CFPB data across replications for varying numbers of topics.}
    \label{tab:cfpb_reliab}
\end{table}

To assess the impact of reliability on downstream analyses, we applied multivariate omega reliability to the CFPB cryptocurrency data using LDA models with 20, 50, and 100 topics. As shown in Table \ref{tab:cfpb_reliab}, reliability significantly decreased with increasing numbers of topics. Similar declines were observed in Table \ref{tab:synth_omega_reliab}, although LDA models were less reliable on the synthetic data. While reliability for 100 topics was within an acceptable range, replications of models with such reliability values are likely to yield diverse results and interpretations. This highlights the importance of careful consideration of reliability when employing topic models, particularly with larger numbers of topics. To illustrate this, we conducted a case study (Appendix \ref{sec:appendix_cfpb_case_study}).

\begin{table}[t]
    \centering
    \begin{tabular}{c|c|c|c|c|c}
         \textbf{Topics} & \textbf{Min} & \textbf{Q1} & \textbf{Q2} & \textbf{Q3} & \textbf{Max} \\
         \hline
         \textbf{20} & 0.67 & 0.73 & 0.75 & 0.77 & 0.81 \\
         \hline
         \textbf{50} & 0.77 & 0.79 & 0.80 & 0.81 & 0.83 \\
         \hline
         \textbf{100} & 0.80 & 0.84 & 0.84 & 0.85 & 0.87
    \end{tabular}
    \caption{Prediction accuracy of logistic regression models across replications when 20, 50, and 100 are used in a CFPB data topic model.}
    \label{tab:cfpb_pred_accuracy}
\end{table}

We develop logistic regression models to predict the binary outcome of a response to a submitted consumer complaint receiving or not receiving a timely response to further explore the downstream analysis impact of deviation across replications. All the topics from each replication's LDA model are used in each replication's prediction model. Table \ref{tab:cfpb_pred_accuracy} summarizes the variation in prediction accuracy across models, with LDA models exhibiting higher predictive power when using more topics. This is expected, as a larger number of topics naturally captures more characteristics of the corpus.

\begin{table}[t]
\begin{subtable}{.45\textwidth}
\centering
    \begin{tabular}{c|c|c|c|c|c}
        \textbf{Topics} & \textbf{Min} & \textbf{Q1} & \textbf{Q2} & \textbf{Q3} & \textbf{Max} \\
        \hline
        \textbf{20} & -3.2& -0.1 & 0.1 & 0.3 & 0.9 \\
        \hline
        \textbf{50} & -19.3 & -3.6 & -0.7 & 0.5 & 2.7 \\
        \hline
        \textbf{100} & -81.7 & -13.8 & -3.8 & 4.6 & 49.8
        
    \end{tabular}
    \caption{Word weightings for the highly prevalent word `transfer.'}
    \label{tab:cfpb_word_weights_transfer}
 \end{subtable}
\begin{subtable}{.45\textwidth}
    \centering
    \begin{tabular}{c|c|c|c|c|c}
        \textbf{Topics} & \textbf{Min} & \textbf{Q1} & \textbf{Q2} & \textbf{Q3} & \textbf{Max} \\
        \hline
        \textbf{20} & -1 & 0 & 0.1 & 0.3 & 0.9 \\
        \hline
        \textbf{50} & -9.8 & 0.4 & 1.3 & 2.3 & 6.7 \\
        \hline
        \textbf{100} & -19.9 & 3.3 & 15 & 25.4 & 42.6
        
    \end{tabular}
    \caption{Word weightings for the highly prevalent word `case.'}
    \label{tab:cfpb_word_weights_case}
\end{subtable}
\caption{Weighting of the two most highly prevalent words across LDA replications for LDA models with varying numbers of topics. All topics are used to predict whether or not the CFPB returns a timely response.}
\label{tab:cfpb_word_weights}
\end{table}

If an LDA model for a given number of topics had high reliability across replications, it would be expected that words would be weighted similarly across replications within the predictive model. We quantify an individual word weighting by multiplying each model coefficient by the respective topic's distribution for the given word, then summing over all products. Table \ref{tab:cfpb_word_weights} describes the variation in word weighting across replications for the two most highly prevalent words that appear consistently in the most prevalent topics for all topics. While the 20 topic model shows some consistency for these high impact words, the 50 and 100 topic models reveals significant variation.
If such a model were used to provide recommendations for developing effective complaint narratives, important guidelines like including a case number could vary dramatically due solely to random initialization. These results align with the model reliability in Table \ref{tab:cfpb_reliab}.

Our findings highlight the essential role of reliability in ensuring interpretability of predictive models that utilize topic models as predictors. Even when prediction accuracy is the primary concern, understanding the underlying factors driving predictions is equally important. Without reliability, insights derived from such predictive models can be compromised, undermining their practical value.

\section{Discussion}
\label{sec:discussion}

Reliability is a critical factor in topic model analysis, significantly influencing subsequent research. Our findings highlight the intricate relationship between topic model complexity, predictive accuracy, and reliability. While increasing the number of topics in an LDA model can enhance feature representation for predictive tasks, it can also compromise the model's reliability when used for corpus-based analysis. Furthermore, inherent randomness in topic models can introduce significant variability, making results less interpretable and reliable.

Our reliability method provides a vital tool that should be a standard component of any topic model-based research, serving as a cornerstone for assessing robustness of topic models and ensuring reliability of subsequent analyses. This method can be extended to alternative topic modeling methods like NMF and BERTopic, a planned direction for our future work \citep{lee1999learning, grootendorst2022bertopic}. Researchers should prioritize reliability as a fundamental component of topic model analysis, akin to the significance of standard errors in statistical modeling. By reporting reliability measures, researchers can present a more comprehensive and accurate understanding of their findings, mitigating the risk of drawing misleading conclusions.

\section{Limitations}

The work in this paper is limited by the existing reliability measurements. Currently, no nonlinear reliability measures exist as all existing measurements rely upon the classical statistical test theory assumptions, of which linearity of errors is included. Given that measurement error in topic models is not expected to be linear, such a measure would strengthen this work. 

While the selected topic matching process facilitates direct comparison to the standard practice method, it simplifies the complex relationship between topics and documents by neglecting document over topic distributions. This simplification, though necessary for comparability, may overlook important aspects of document over topic associations. Future research could explore more comprehensive topic matching methods that explicitly consider the distribution of documents over topics and alternative similarity measures that incorporate additional semantic information. It's important to note that the selected topic matching method was primarily driven by comparability with existing standards, and future research could investigate the impact of different topic matching methods on the proposed approach's overall performance.

The \citet{maier2018applying} method has emerged as the de facto standard for topic model reliability assessment, evidenced by its widespread adoption and significant research attention, serving as a robust and representative benchmark for stability assessment and providing a solid foundation for comparison. We emphasize that similarity measures are not a measure of reliability and are not statistically grounded, a core weakness of the \citet{maier2018applying} methodology. Following this reasoning, we argue that all related methods employing similarity measures are similarly weakened by the lack of statistical theory to back up the method and assume the limitations accordingly. Despite the practical relevance of the standard practice baseline and its robust and representative nature as a benchmark for stability assessment, comparing to a single baseline presents a limitation in scope. Future work could investigate the impact of different baselines on our proposed reliability measure and explore the potential benefits of incorporating alternative approaches.

The work in this paper is limited to LDA topic models, but provides a framework that can be extended to alternative topic modeling methods like NMF and BERTopic \citep{lee1999learning, grootendorst2022bertopic}. As we note in Section \ref{sec:discussion}, this is a planned direction for future work.

\section{Ethics Statement}

No ethical issues are posed by the theory  or application discussed in this work. The application data from the CFPB public website is anonymized and contains no unique identifiers, thereby preventing any potential privacy violations. Future applications of the methods developed in this paper may pose ethical issues depending on the proprietary nature of future data sets. Required privacy regulations should be followed closely by researchers applying these methods to proprietary datasets.  

\bibliography{custom}

\clearpage

\appendix

\section{Detriment of Standard Practice Cutoff}
\label{sec:appendix_std_prac_cutoff}

\begin{figure}[!ht]
\centering
\includegraphics[width=\columnwidth]{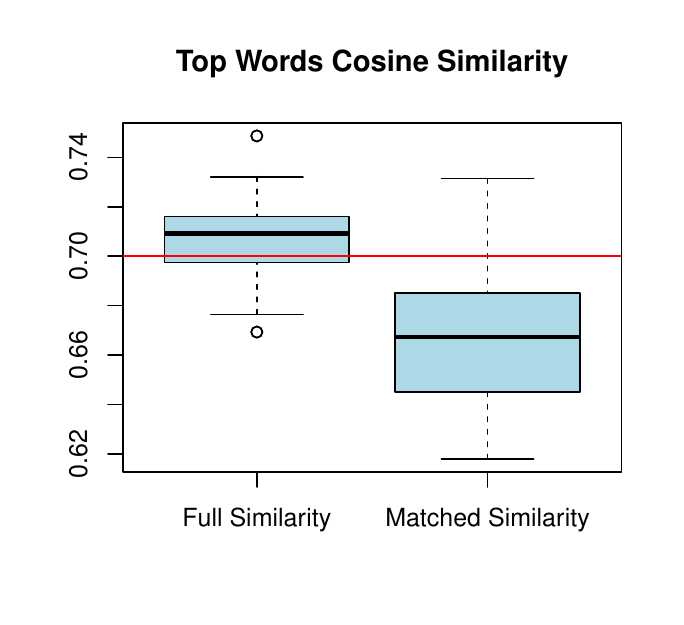}
\caption{Maximal cosine similarity for top words distributions for each topic. In the `Full' case, all other topics are considered for the maximal similarity measure. In the `Matched' case, topics are matched with best available topic and cosine similarity is determined for each pairing.}
\label{fig:cossim_n100}
\end{figure}

To properly align topics, the maximal cosine similarity of the top words distribution for each topic is used. All topics must be matched with one singular other topic in practice, however, so the maximal cosine similarity is somewhat lower in this setting. Figure \ref{fig:cossim_n100} depicts this difference, with cosine similarity for top words in the matched case being slightly less than that of unmatched topics. Using the hard cutoff of 0.7, the distribution shifts slightly, but the reliability value for the \citet{maier2018applying} reliability measure changes drastically as the proportion of values above and below the 0.7 cutoff is extremely different for the full and matched cases. Clearly, the hard cutoff does not allow for adequate characterization of the distribution of cosine similarities, presenting a major issue for the current standard practice measurement.

\section{Model Selection}
\label{sec:appendix_mod_selction}

\begin{figure}[t]
\centering
\includegraphics[width=\columnwidth]{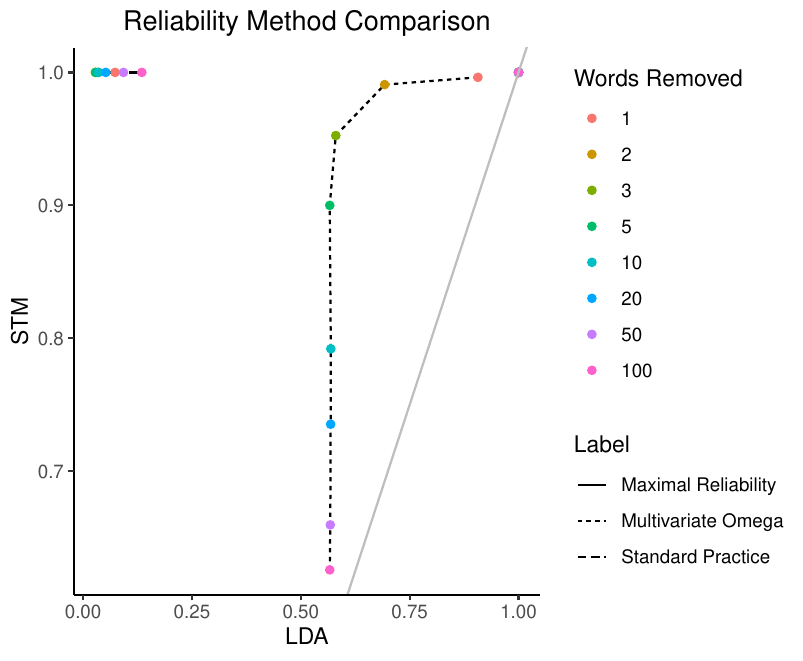}
\caption{Comparison of reliability methods of \texttt{stm} on LDA under varying numbers of random vocabulary word removals. The grey line here depicts the y=x relation.}
\label{fig:method_comp_stm_lda}
\end{figure}

Given the impact of word removal on topic models, we would expect the reliability of \texttt{stm} on LDA to follow along the y=x line given that \texttt{stm} is deterministic and the LDA initialization seed is held constant. Figure \ref{fig:method_comp_stm_lda} visualizes this comparison. As expected, the standard practice method deviates drastically from the y=x relation. This makes sense given its previously discussed poor characterization of the reliability. Both multivariate omega and maximal reliability fall more closely along the y=x line. For example, in the 1 word removal setting for a randomly selected topic, consider the top 10 documents containing the highest proportions of the topic. Across replications, \texttt{stm} has 7 of the same documents in the top 10 in 9 different replications while LDA has none of the same documents within the top 10 across more than 2 replications. Meanwhile, both \texttt{stm} and LDA possess strong consistencies within the vocabulary over topic matrices across replications as seen by strong alignment of most prevalent words in the most prevalent topics across replications. From these results, then, we recommend multivariate omega as the optimal reliability measure.   

\begin{table}[t]
    \begin{subtable}{.45\textwidth}
 \centering
\begin{tabular}{c|c|c|c}
 & \textbf{Word 1} & \textbf{Word 2}  & \textbf{Word 3} \\
 \hline
\textbf{Rep 1} & peculiar & spirit & sort \\ 
\hline
\textbf{Rep 2} & peculiar & spirit & economi \\ 
\hline
\textbf{Rep 3} & peculiar & spirit & sort \\ 
\hline
\textbf{Rep 4} & peculiar & spirit & sort \\ 
\hline
\textbf{Rep 5} & peculiar & spirit & sort \\ 
\hline
\textbf{Rep 6} & peculiar & spirit & sort \\ 
\hline
\textbf{Rep 7} & peculiar & spirit & consult \\ 
\hline
\textbf{Rep 8} & peculiar & spirit & consult \\ 
\hline
\textbf{Rep 9} & peculiar & spirit & communic \\ 
\hline
\textbf{Rep 10} & peculiar & spirit & terror \\ 
\end{tabular}
\caption{1 word removed}
\label{tab:frex_stm_1rm}
\end{subtable}

 \begin{subtable}{.45\textwidth}
 \centering
\begin{tabular}{c|c|c|c}
 & \textbf{Word 1} & \textbf{Word 2}  & \textbf{Word 3} \\
 \hline
\textbf{Rep 1} & peculiar & pari & communic \\ 
\hline
\textbf{Rep 2} & communic & peculiar & sort \\ 
\hline
\textbf{Rep 3} & peculiar & sort & spirit \\ 
\hline
\textbf{Rep 4} & peculiar & communic & shane \\ 
\hline
\textbf{Rep 5} & will & may & alexa \\ 
\hline
\textbf{Rep 6} & may & alexa & will \\ 
\hline
\textbf{Rep 7} & spirit & peculiar & nds \\ 
\hline
\textbf{Rep 8} & peculiar & blu & exc \\ 
\hline
\textbf{Rep 9} & friday & agen & walsh \\ 
\hline
\textbf{Rep 10} & will & may & fame \\ 
\end{tabular}
\caption{100 words removed}
\label{tab:frex_stm_100rm}
\end{subtable}
\caption{Top three frequent and exclusive words from the most prevalent topic within each replication when one word (Subtable \subref{tab:frex_stm_1rm}) and 100 words (Subtable \subref{tab:frex_stm_100rm}) are removed at random from each replication.}
%\subref{tab:frex_stm_100rm}
\label{tab:frex_stm_both}
\end{table}

To further investigate the results of the \texttt{stm} models with varying numbers of words removed, we consider the frequent and exclusive (FREX) words of the most prevalent topics described in Table \ref{tab:frex_stm_both}. When only one word is removed at random from the corpus, as in Table \ref{tab:frex_stm_1rm}, the top two FREX words are identical across replications. Five of the replications are identical across the top 3 FREX words as well, with deviations present in the remaining 5 replications. Drastic deviations, however, are present when 100 words are removed at random from the corpus, with minimal (if any, depending on the replication) consistency across replications, as showcased in Table \ref{tab:frex_stm_100rm}. This maps directly to the performance of the multivariate omega measure, as shown in Figures \ref{fig:method_comp_stm} and \ref{fig:method_comp_stm_lda}.

\section{Application Case Study}
\label{sec:appendix_cfpb_case_study}

    \begin{table}[t]
    \begin{subtable}{.45\textwidth}
 \centering
\begin{tabular}{c|c|c}
    \textbf{Topic 12} & \textbf{Topic 14} & \textbf{Topic 19} \\
   \hline
    coinbas & financi & money \\
    account & reason & get \\
    access & coinbas & account 
    
\end{tabular}
\caption{Replication 1}
\end{subtable}

 \begin{subtable}{.45\textwidth}
 \centering
\begin{tabular}{c|c|c}
    \textbf{Topic 3} & \textbf{Topic 6} & \textbf{Topic 11} \\
   \hline
    transact & account & xxxx \\
    payment & verifi & financi \\
    pend & inform & coinbas 
    
\end{tabular}
\caption{Replication 2}
\end{subtable}
\caption{Example comparison of 3 most prevalent words of 3 most prevalent topics for two of the replications with 20 topics.}
\label{tab:cfpb20_reps}
\end{table}

       \begin{table}[t]
       \begin{subtable}{.45\textwidth}
 \centering
\begin{tabular}{c|c|c}
    \textbf{Topic 10} & \textbf{Topic 20} & \textbf{Topic 24} \\
   \hline
    person & sell & card \\
    name & trade & purchas \\
    inform & buy & credit 
    
\end{tabular}
\caption{Replication 1}
\end{subtable}

 \begin{subtable}{.45\textwidth}
 \centering
\begin{tabular}{c|c|c}
    \textbf{Topic 9} & \textbf{Topic 42} & \textbf{Topic 48} \\
   \hline
    state & transact & email \\
    messag & fee & receiv \\
    due & charg & respons 
    
\end{tabular}
\caption{Replication 2}
\end{subtable}
\caption{Example comparison of 3 most prevalent words of 3 most prevalent topics for two of the replications with 50 topics.}
\label{tab:cfpb50_reps}
\end{table}

Tables \ref{tab:cfpb20_reps} and \ref{tab:cfpb50_reps} display a comparison of the three most prevalent words from each of the three most prevalent topics for from example replications for 20 and 50 topics respectively. In Table \ref{tab:cfpb20_reps}, topics 14 and 11 have two words in common and seem to describe similar topics, topics 19 and 6 have one of the three top words in common, and topic 3 has no commonalities with replication 1. Given these shared words, it is evident that there is some, though minimal, consistency across replications, but not enough to warrant any clear consistencies through the replications. The reliability value of 0.644 not only makes sense in this context, but is also fitting. In the 50 topics case, as described in Table \ref{tab:cfpb50_reps}, none of the words align across replications. Here, then, we can see that a reliability value as low as 0.386 (significantly below the unacceptable threshold) is easily plausible. 

\end{document}